\documentclass{article} 
\usepackage{iclr2021_conference,times}

\usepackage{amsmath,amsfonts,bm}

\def\Figref#1{Figure~\ref{#1}}

\def\eqref#1{equation~\ref{#1}}
\def\Eqref#1{Equation~\ref{#1}}

\def\1{\bm{1}}

\def\v_#1{{\bm #1}}

\def\vh{{\bm{h}}}

\def\vs{{\bm{s}}}

\def\vu{{\bm{u}}}
\def\vv{{\bm{v}}}

\def\evv{{v}}

\DeclareMathAlphabet{\mathsfit}{\encodingdefault}{\sfdefault}{m}{sl}
\SetMathAlphabet{\mathsfit}{bold}{\encodingdefault}{\sfdefault}{bx}{n}

\newcommand{\softmax}{\mathrm{softmax}}
\newcommand{\sigmoid}{\sigma}

\usepackage[pdfusetitle]{hyperref}
\usepackage{url}
\usepackage{subcaption}
\usepackage{graphicx}
\usepackage[labelfont=bf]{caption}
\usepackage{multicol}
\usepackage{siunitx}
\usepackage{float}

\hypersetup{
    colorlinks=true,
    linkcolor=red,
    citecolor=blue,
    filecolor=magenta,
    urlcolor=cyan,
}

\title{Deep Spiking Neural Networks \\ 
with Resonate-and-Fire Neurons}

\makeatletter
\def\blfootnote{\gdef\@thefnmark{}\@footnotetext}
\makeatother

\author{Badr AlKhamissi \\
Independent \\
\texttt{badr [at] khamissi.com} \\
\And
Muhammad ElNokrashy \\
Microsoft EGDC \\
\texttt{muhammad.nael [at] gmail.com} \\
\And
D. Bernal-Casas \\
U Barcelona, Spain \\
\texttt{} \\
}

\iclrfinalcopy 
\begin{document}

\maketitle

\begin{abstract}


In this work, we explore a new \emph{Spiking Neural Network} (SNN) formulation with \emph{Resonate-and-Fire} (RAF) neurons \citep{Izhikevich_2001} trained with gradient descent via back-propagation. The RAF-SNN, while more biologically plausible, achieves performance comparable to or higher than conventional models in the Machine Learning literature across different network configurations, using similar or fewer parameters. Strikingly, the RAF-SNN proves robust against noise induced at testing/training time, under both \emph{static} and \emph{dynamic} conditions. Against CNN on MNIST, we show $25\%$ higher absolute accuracy with $\mathcal{N}(0,0.2)$ induced noise at testing time. Against LSTM on N-MNIST, we show $70\%$ higher absolute accuracy with $20\%$ induced noise at training time.



\end{abstract}

\section{Introduction}


Artificial intelligence traces many of its roots back to research in neuroscience and psychology---research seeking to understand the human brain, from the neuronal to the behavioral level. For example, Convolutional Neural Networks (CNNs) were inspired by the hierarchical feed-forward structure of the visual cortex \citep{hubelwiesel68, fukushimaneocognitron80}. Reinforcement Learning (RL) branched out from psychology research on animal conditioning \citep{rescorla72}. Interestingly, there are many instances where AI progress has inspired new brain theories that were later empirically verified. For example, RL research inspired a reward-based learning theory of dopaminergic function \citep{schultz97}. While deep CNNs trained on natural images were shown to reproduce neurophysiological patterns observed in animals \citep{Lindsay_2020}. Both fields have mutually benefited one another in what has been recently called ``a virtuous circle'' \citep{hassabis17}.


In this light, the scientific community has been exploring Spiking Neural Networks (SNNs). SNNs employ simplified models that approximate neuronal mechanisms we believe the brain uses to process discrete spatio-temporal events (the spikes). One prominent such model is the \emph{Leaky-Integrate-and-Fire} (LIF) neuron.
In LIF, the neuron integrates the inputs over time, firing when the \emph{potential} passes a set threshold.
Hardware architectures have been developed to exploit this event-based behaviour \citep{Merolla14, ankit2017resparc, davies2018loihi}. Their results are auspicious for achieving ultra-low power processing of event-based data streams. For example, in deep learning architectures with spiking neurons, it was observed that the number of spikes drops significantly at deeper layers, reducing the computation requirements for neuromorphic hardware \citep{rueckauer2017conversion, sengupta2019going}. A caveat of these approaches is that the neuron model utilized (the LIF neuron) can not reproduce relevant features of cortical neurons \citep{Izhikevich_2001}.

Towards overcoming this limitation, we consider a novel SNN with a more biologically plausible neuron model: the \emph{Resonate-and-Fire} (RAF) model \citep{Izhikevich_2001}. The RAF neuron can model more neurodynamics given its characteristic as a resonator model, and research has shown it to replicate biological neural data, despite its simplicity compared to other resonator neuronal models \citep{Torikai_2009, Pauley_2018}. 

\section{Methods and Experiments}

\begin{figure}[H]
\begin{subfigure}{0.6\textwidth}
    \centering
    \includegraphics[width=1.0\linewidth]{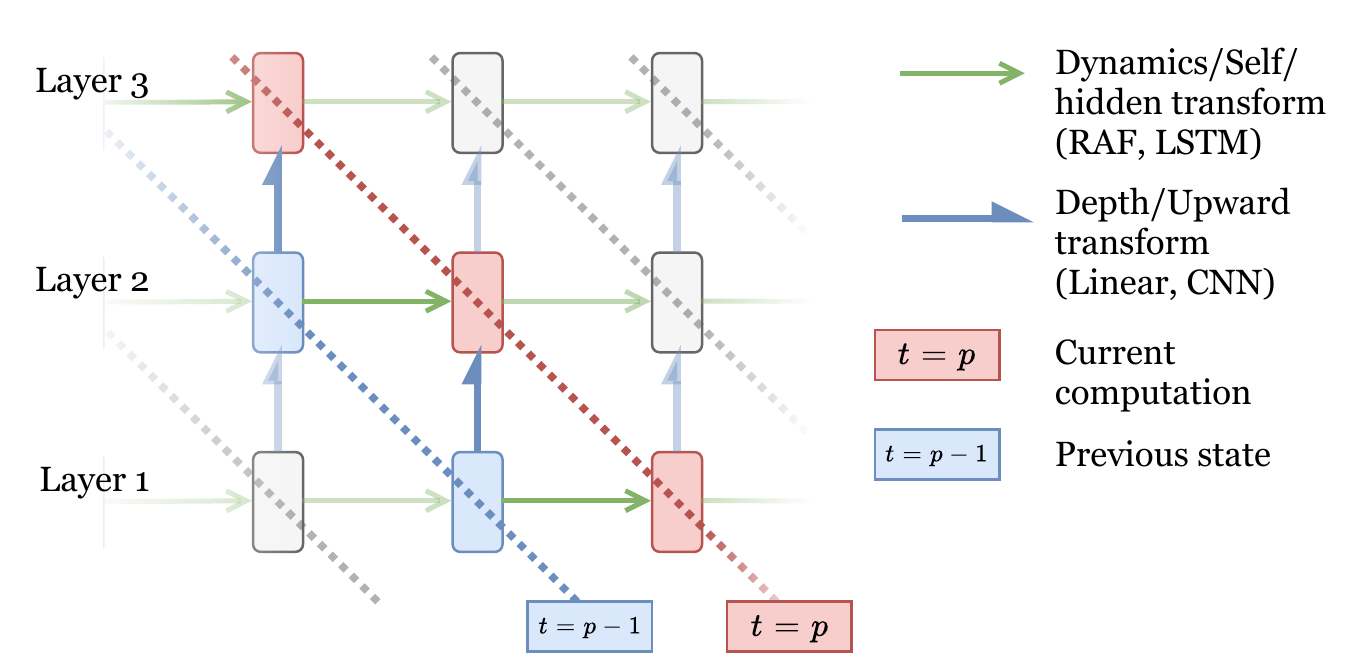}
    \caption{}
    \label{fig:forward_pass}
\end{subfigure}%
\begin{subfigure}{0.4\textwidth}
    \centering
    \includegraphics[width=1.0\linewidth]{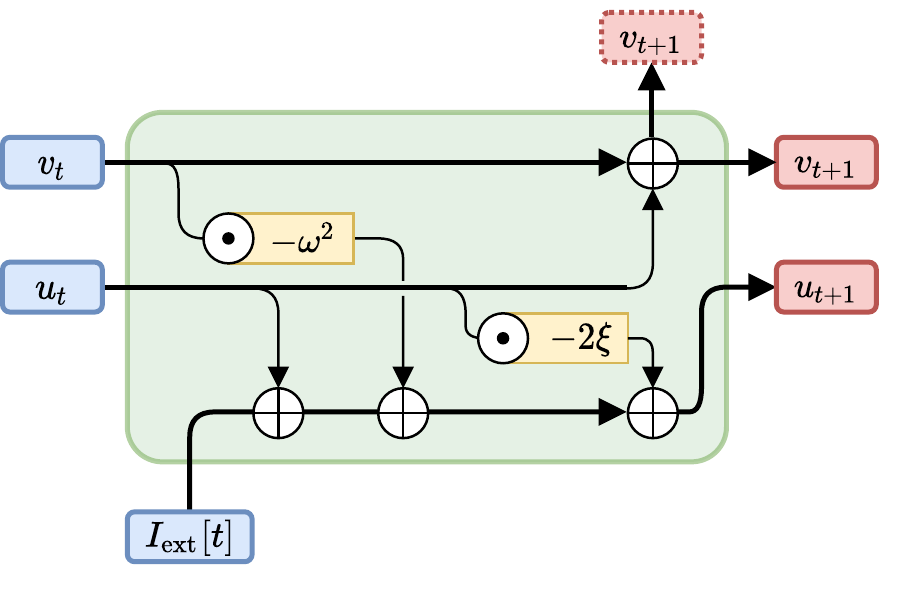}
    \caption{}
    \label{fig:cell}
\end{subfigure}
\caption{\textbf{(a)} The forward pass of the (Upward)\protect\footnotemark \,RAF-SNN, unrolled in time. \textbf{(b)} The RAF\textsuperscript{\ref{foot:upward_raf}} neuron in the style of conventional recurrent cells such as the LSTM.}
\end{figure}

\footnotetext{\label{foot:upward_raf}We consider RAF-SNNs with only 1 kind of inter-neuron connections: From neurons in layer $\ell$ to neurons in layer $\ell+1$. Biologically, this is not a known constraint.}

Figure \ref{fig:forward_pass} visualizes an Upward\textsuperscript{\ref{foot:upward_raf}} RAF-SNN unrolled in time. A time unit $dt$ is defined as a single step forward for each cell in the network. Thus, notice that input at $t=0$ propagates through the network such that the corresponding output would be at $t=L-1$, for depth $L$, not at $t=0$. This formulation is useful for dealing with the dynamical system of the RAF neuron, but is otherwise functionally equivalent to conventional RNN cells, like the LSTM, with the caveat that observed outputs are shifted in time by $+L-1$. The self-transforms $\vv^\ell_{t-1} \mapsto \vv^\ell_{t}$ (green) correspond to $\vh^\ell_{\tau-1} \mapsto \vh^\ell_{\tau}$. The depth-transforms $\vv^{\ell-1}_{t-1} \mapsto \vv^{\ell}_{t}$ (blue) correspond to $\vh^{\ell-1}_{\tau} \mapsto \vh^\ell_{\tau}$, and so on.



\subsection{The RAF Neuron Model}

\Eqref{eq:raf_2d} shows the RAF neuron model. It describes the dynamics of the membrane potential of a neuron with the equation of motion of a forced, damped harmonic oscillator:
\begin{align}
	\frac{d^2 v}{dt^2} + 
    2 \xi \frac{d v}{dt} + 
	\omega^2 v
	&=
	I_\text{ext}[t] 
	\label{eq:raf_2d}
\end{align}
\begin{align}
    V_t &\approx V_{t-1} + dt \odot u_{t-1} , \\
    u_t &\approx u_{t-1} + dt \odot \left(I_\text{ext} - 2\xi u_{t-1} - \omega^2 V_{t-1} \right).
	\dot V &= u_{t-1} , \\
	\dot u &= I_\text{ext} + -\omega^2 V_{t-1} - 2\xi u_{t-1} , \\
	x_{t} &= x_{t-1} + \dot x dt \quad \forall x \in \{u, V\}
	.
\end{align}

where $v$ is the membrane potential, $\xi$ is the damping factor, $\omega$ is the natural frequency, and $I_\text{ext}$ is any external current to the neuron (i.e., the weighted summation of pre-synaptic spikes). 





\subsection{Datasets}

We evaluate our formulation on both the static and the dynamic (or neuromorphic) versions of the MNIST dataset \citep{lecun1998gradient, Orchard15nmnist}. The static version comprises $28 \times 28$ gray-scale images of hand written digits, while the neuromorphic version (``N-MNIST'') was derived from the static dataset by panning and tilting a Dynamic Vision Sensor (DVS) \citep{Lichtsteiner08DVS} in front of a screen displaying the digits. Each sample consists of a \SI{300}{\milli\second} period of ON and OFF events that represent increases or decreases in pixel intensity. The data-stream is pre-processed to be a sequence of $34 \times 34 \times 2$ tensors, similar to \cite{lee16training}, with a sampling time of \SI{10}{\milli\second}.



\subsection{Experimental Setup}
\label{sec:experimental_setup}

\paragraph{Models} We consider $4$ models: RAF vs.\ LSTM on N-MNIST, and RAF-CNN vs.\ CNN on MNIST.
RAF models follow the recurrence in Equations \ref{eq:system_1}, \ref{eq:system_2} \& \ref{eq:system_3}. Beyond that, the RAF-CNN substitutes a convolutional layer for the linear at $f^{(\ell)}(s)$. Architecture parameters (e.g.\ depth and width) are matched for LSTM \& RAF networks, and CNN \& RAF-CNN.
Further details in Appendix \ref{app:arch}.

\paragraph{Poisson Encoding} For the RAF-CNN model, the static inputs are first encoded as Poisson-distributed spike trains where the intensity of each pixel defines the event rate---a standard practice for SNNs \citep{oconnor13, diehl15}.   

\paragraph{Perturbing Images} In the \emph{static} case, we add Gaussian noise of a certain std-dev to the pixel values before Poisson Encoding. In the \emph{dynamic} case, we flip individual binary pixels with some probability $p$ at every time step.

\paragraph{Optimization} Since SNNs deal with  discontinuous spikes (binary activations), we employ the method proposed by \cite{yanguas2020coarse} to enable training with back-propagation. The forward pass follows Equation \ref{eq:hardsoft_forward}, while the gradients are based on the smooth approximation in \Eqref{eq:hardsoft_back}:
\begin{align}
    s^{{\ell}j}_t &= H(v^{{\ell}j}_t - \theta^{{\ell}j}),
    \label{eq:hardsoft_forward} \\
    \partial~s^{{\ell}j}_t &= \partial~\sigmoid(\beta^{{\ell}j}(v^{{\ell}j}_t - \theta^{{\ell}j})).
    \label{eq:hardsoft_back}
\end{align}
where $s^{{\ell}j}_t$ is the output spike of neuron $j$ in layer $\ell$ at time $t$, $v^{{\ell}j}_t$ is its membrane potential, $\theta^{{\ell}j}$ is the firing threshold, $\beta^{{\ell}j}$ is a regularization parameter (controls the steepness of the sigmoid approximation), $H(\cdot)$ is the Heaviside function, and $\sigmoid(\cdot)$ is the sigmoid function. Per neuron, $\{\omega, \xi, \theta, \beta\}$ are learnable. Equations \ref{eq:system_1} \& \ref{eq:system_2} are first-order approximations of \Eqref{eq:raf_2d} in vector form. $f^{(\ell)}(\cdot)$ is a learned forward transform, such as a linear or convolutional layer.
\begin{align}
    \vv_{t+1} &= \vv_{t}
        +
        dt
        \odot
        \vu_{t}
    \label{eq:system_1}
    , \\
    \vu_{t+1} &= \vu_{t}
        +
        dt
        \odot
        \left({\bm I}_\text{ext}[t] - 2{\bm\xi}\odot \vu_{t} - {\bm\omega}^2 \odot\vv_{t} \right) 
    \label{eq:system_2}
    , \\
    {\bm I}_\text{ext}^{\ell}[t] &= f^{(\ell)}( \vs_{t}^{\ell-1})
    \label{eq:system_3}
    .
\end{align}
We use the AdamW optimizer \citep{adamw} with an initial learning rate of $0.01$. The model is trained until the validation accuracy fails to improve for 6 consecutive epochs, where each epoch enumerates the training set in random order. The learning rate is halved when the validation loss does not improve for $2$ epochs. The mini-batch size is $32$. Neuron parameters are initialized by uniform sampling: $\omega \sim \mathcal{U}(0, 1.1 \times 2\pi)$, $\xi \sim \mathcal{U}(0, 2.5)$, $\theta \sim \mathcal{U}(0, 2.5)$. While $\beta_\text{init} = 5$.
\paragraph{Objective Function} In the static case, optimizing the cross-entropy loss on the last observed membrane potential $\evv_{t=\text{last}}$ yields the best results. In the dynamic case, we compute the cross-entropy loss on the number of spikes in the last $15$ time steps as values to the $\softmax$. 





\section{Results}
\label{sec:results}

\def\footbaseline{\textsuperscript{\ref{foot:graph_baseline_indicator}}}
\begin{figure}[H]
\centering
\begin{subfigure}{.33\textwidth}
    \centering
    \includegraphics[width=1\linewidth]{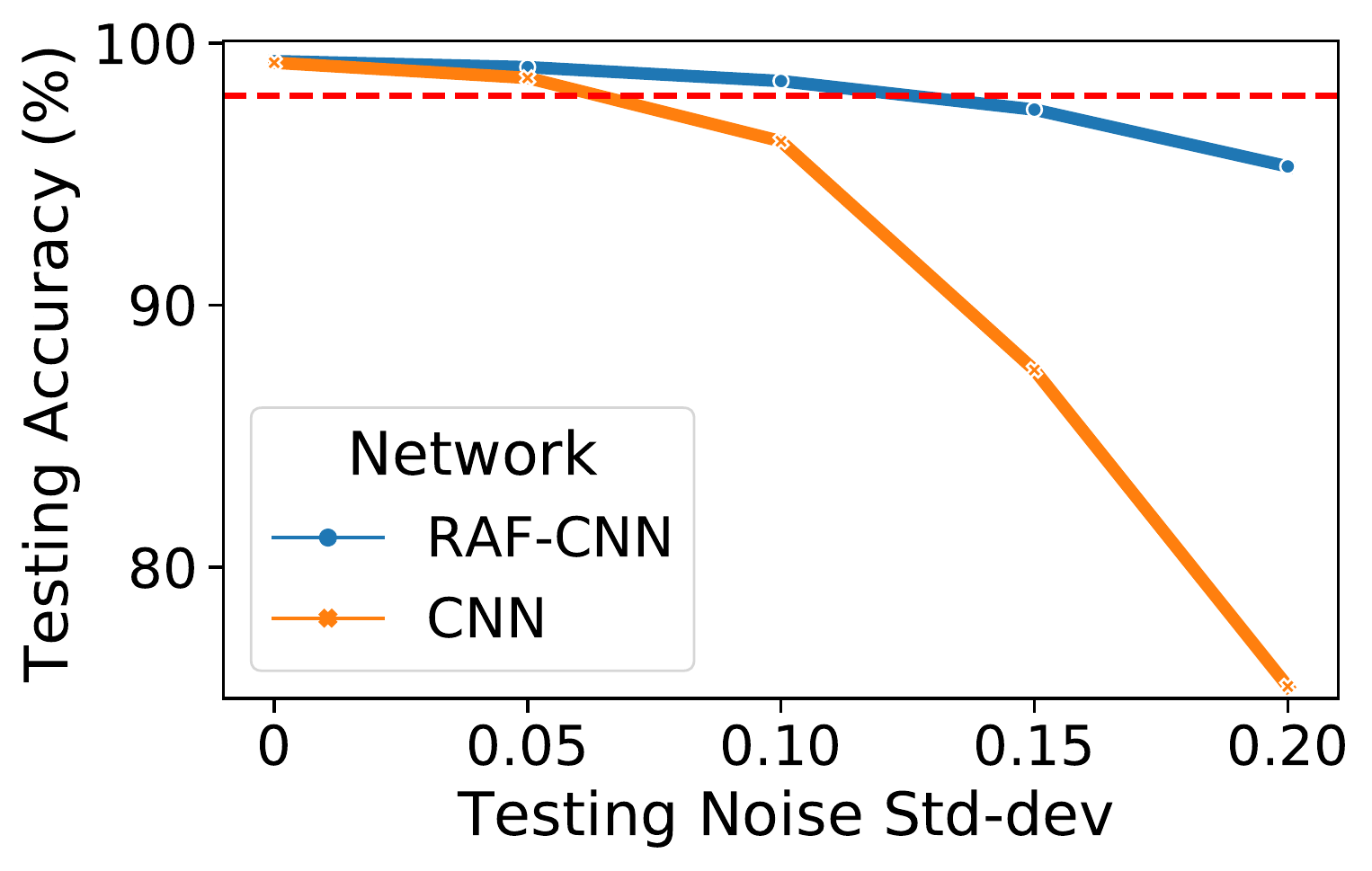}
    \caption{} 
    \label{fig:static_test_noise}
\end{subfigure}%
\begin{subfigure}{.33\textwidth}
    \centering
    \includegraphics[width=1\linewidth]{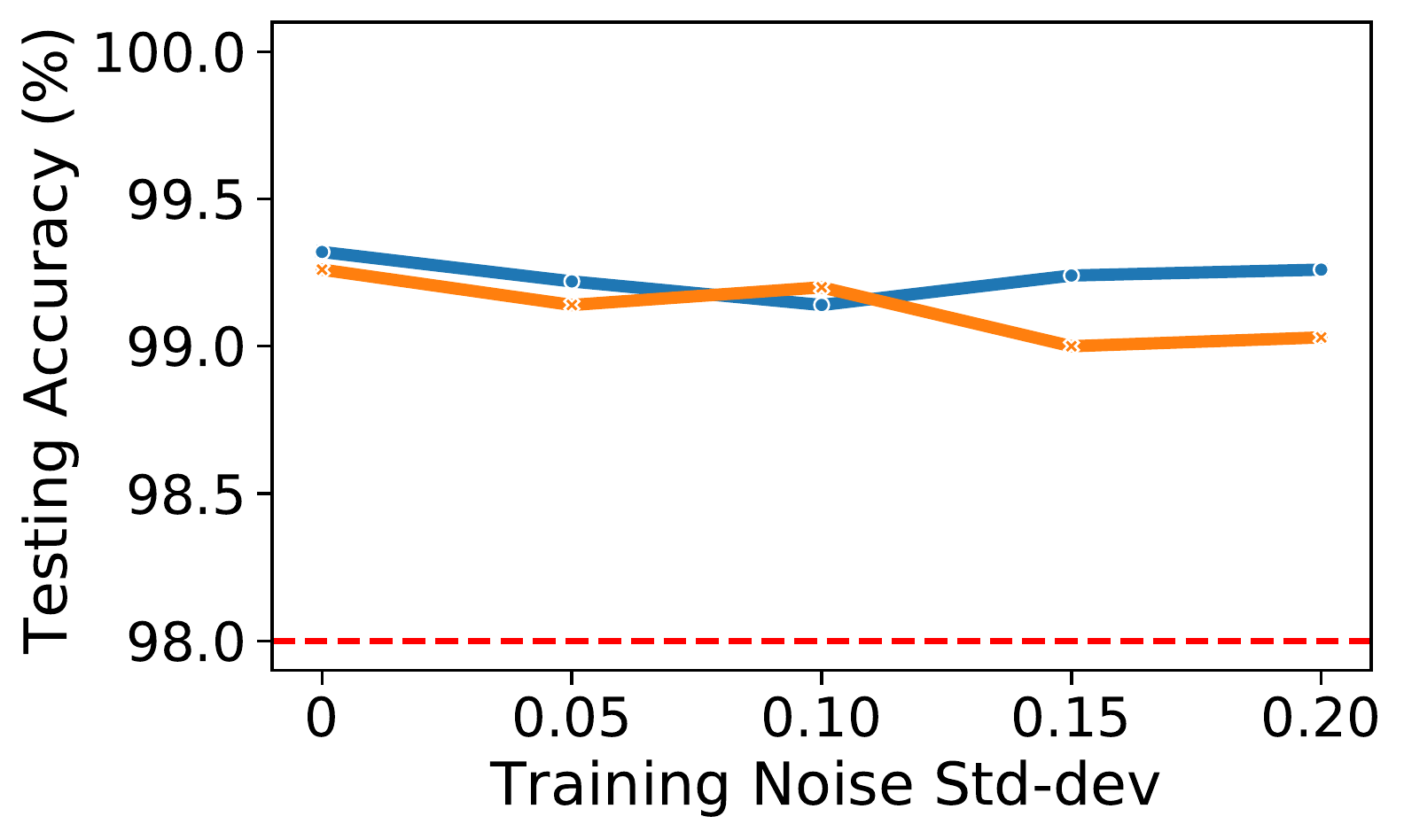}
    \caption{} 
\end{subfigure}%
\begin{subfigure}{.33\textwidth}
    \centering
    \includegraphics[width=1\linewidth]{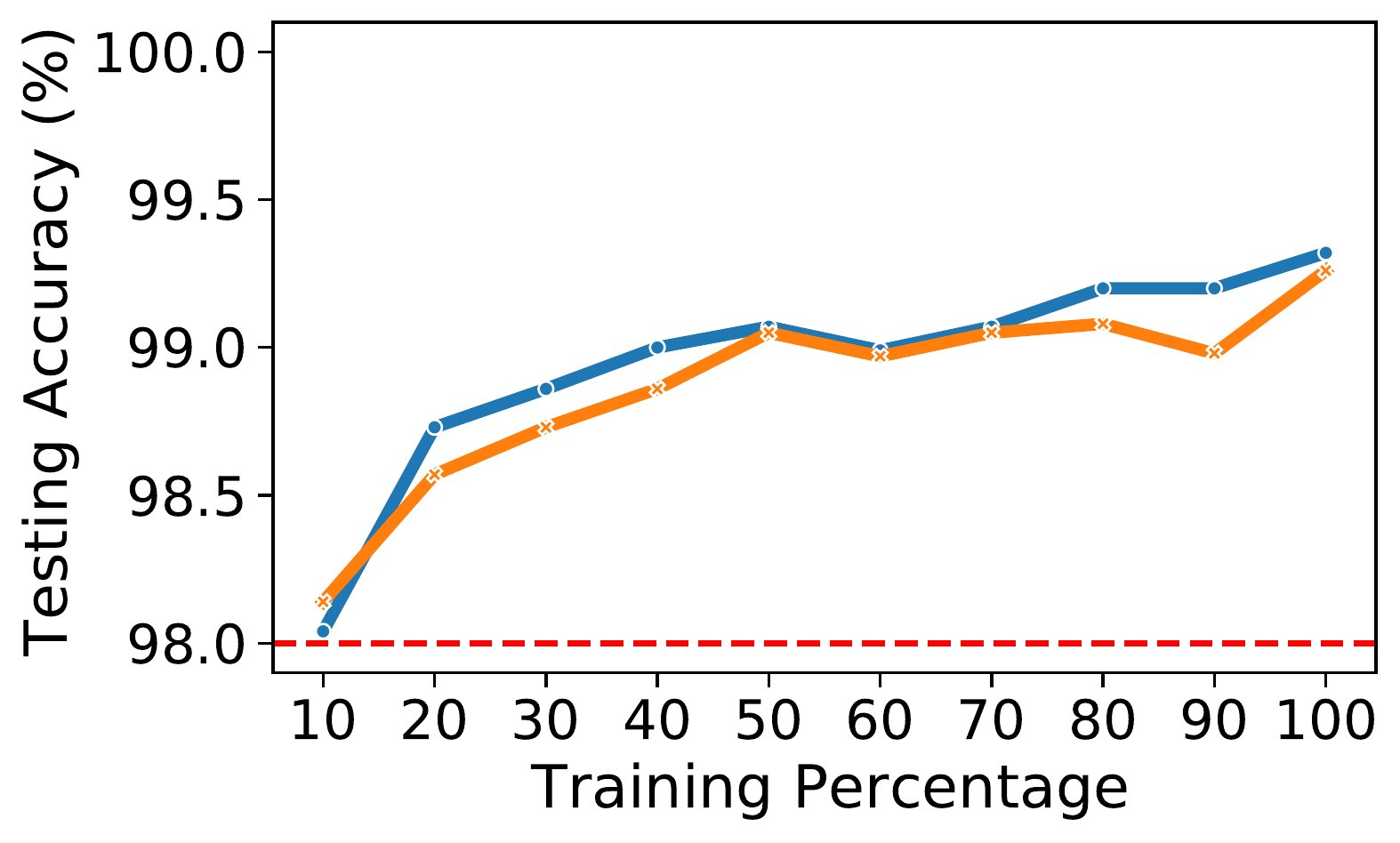}
    \caption{} 
\end{subfigure}
\begin{subfigure}{.33\textwidth}
    \centering
    \includegraphics[width=1\linewidth]{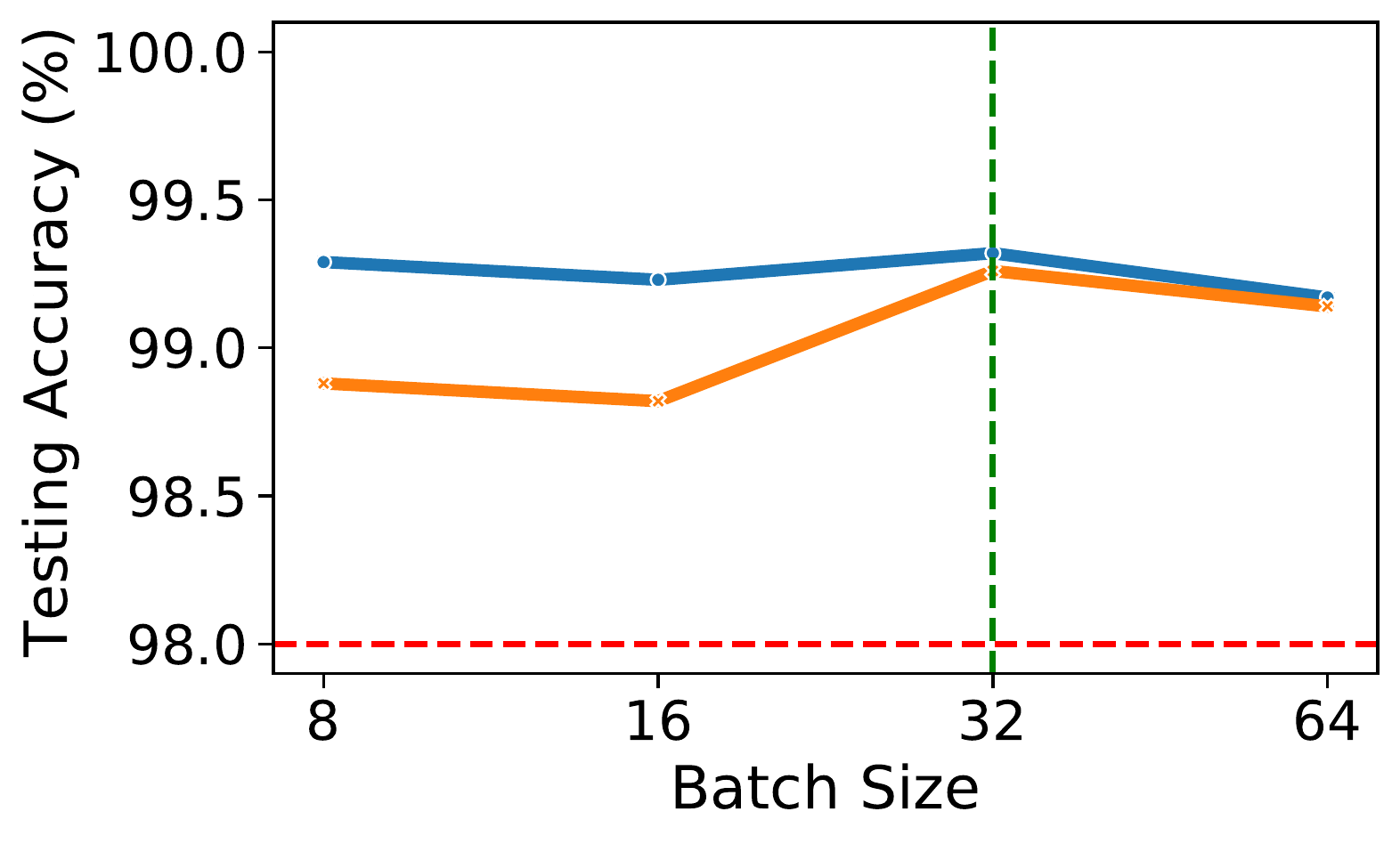}
    \caption{} 
\end{subfigure}%
\begin{subfigure}{.33\textwidth}
    \centering
    \includegraphics[width=1\linewidth]{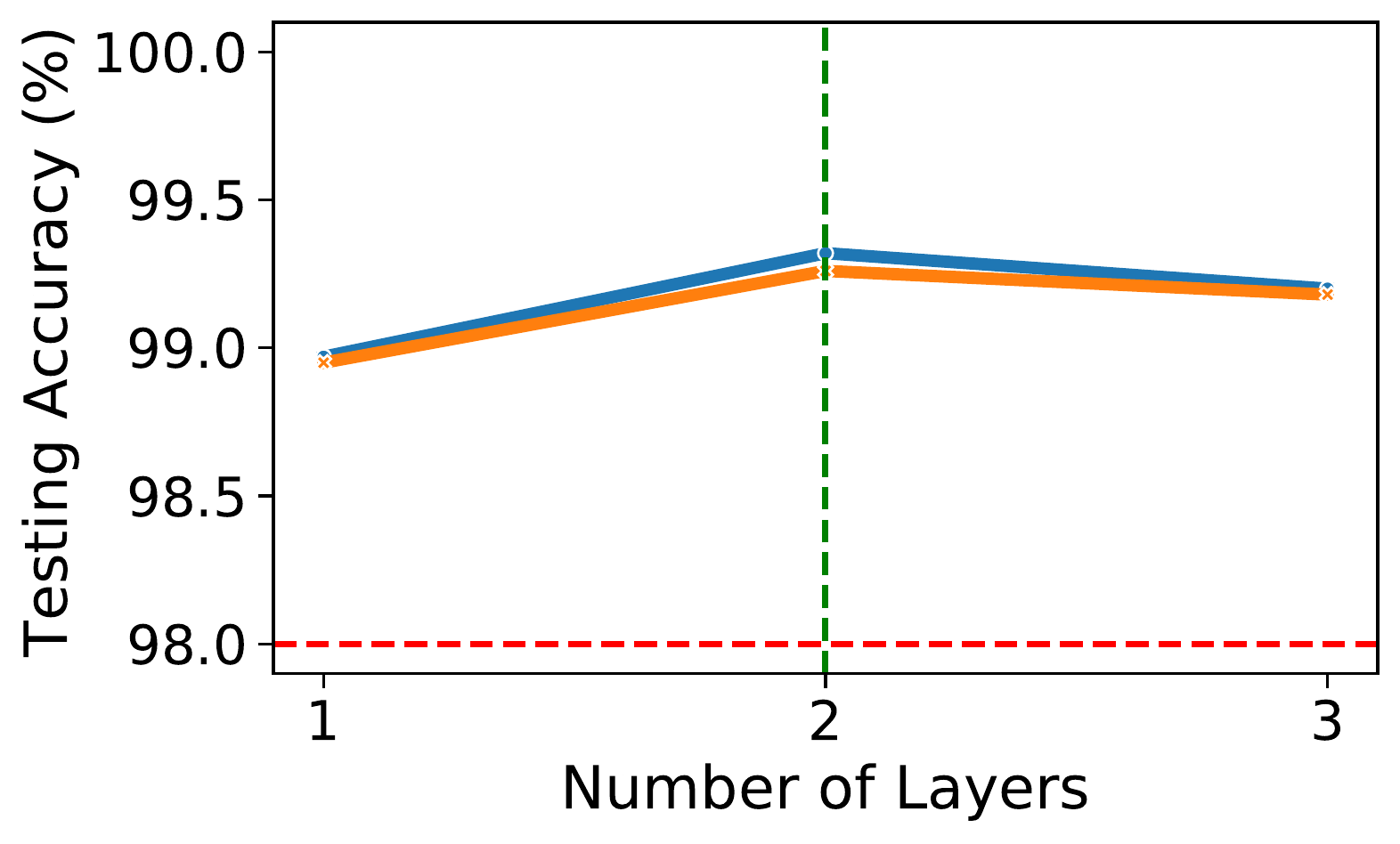}
    \caption{} 
\end{subfigure}%
\begin{subfigure}{.33\textwidth}
    \centering
    \includegraphics[width=1\linewidth]{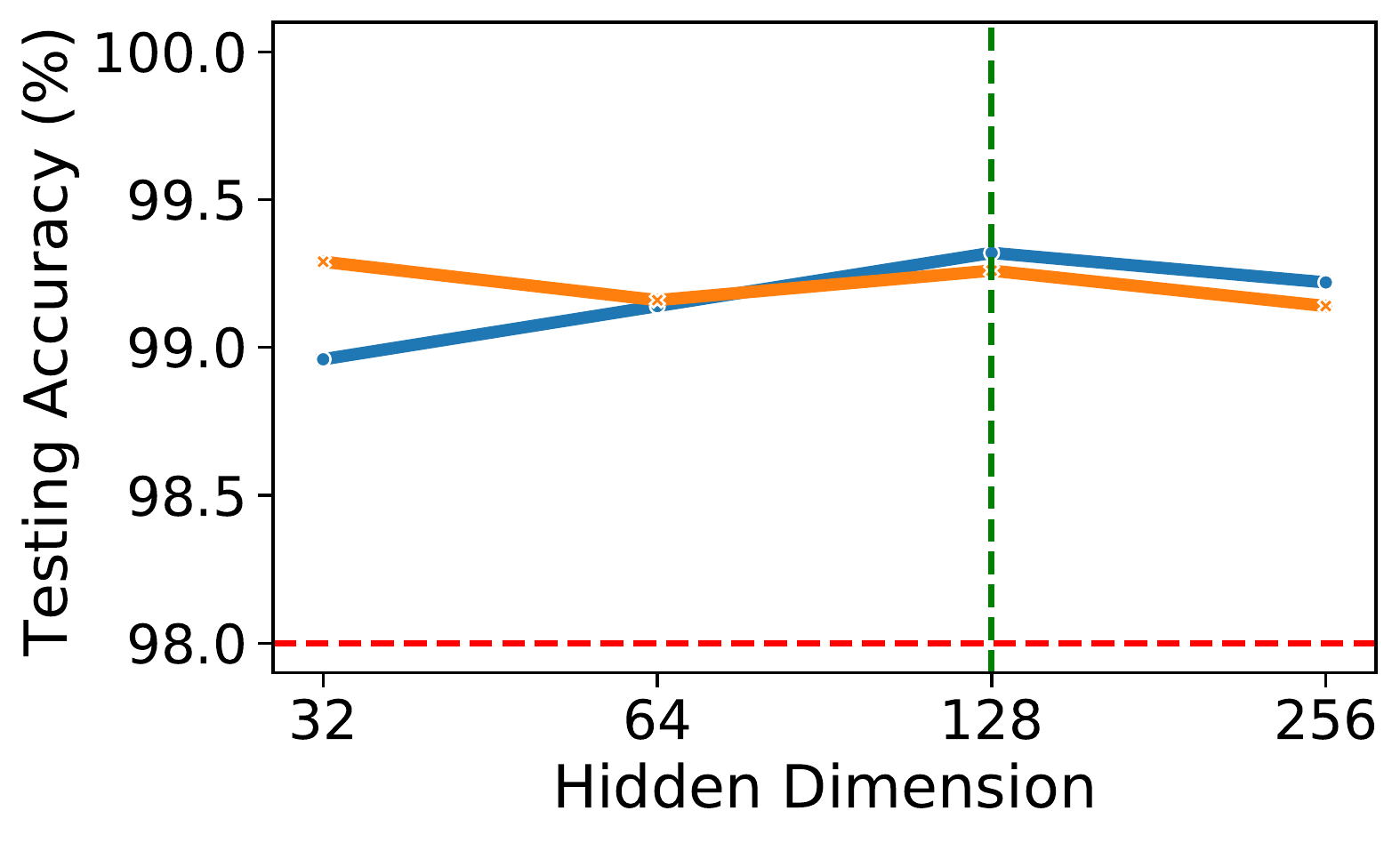}
    \caption{} 
\end{subfigure}

\caption{\textbf{Test accuracy on the MNIST dataset (static).} Comparison of RAF-CNNs vs.\ CNNs as a function of: \textbf{(a)} Noise std-dev at test, \textbf{(b)} Noise std-dev at train, \textbf{(c)} Training set size, \textbf{(d)} Batch size\footbaseline, \textbf{(e)} Depth\footbaseline, and \textbf{(f)} Width\footbaseline. The red line is a visual anchor at $98\%$ accuracy.
}
\label{fig:static_figures}
\end{figure}

\small{* Each data point is the average of $8$ runs, to account for the probabilistic noise and the Poisson encoder.}



In the static (or synthesized) case (\Figref{fig:static_figures}), we observe model behavior when adding Gaussian noise to the \textit{base} input images across a range of std-dev values. With up to $\sigma\!=\!0.2$ noise added at \textit{testing} time for a network trained with clean input: The RAF-CNN maintains performance, while the CNN degrades by up to $25\%$ accuracy. Both networks, however, maintain performance when fed increasingly noisy inputs at \textit{training} time, as measured on the clean test set. Both networks also present similar performance trajectories when varying the training data size, although the RAF-CNN manages to lead very slightly. In addition, we tested across a range of values for batch size, network width, and depth---both models behaved similarly.


\begin{figure}[h]
\centering
\begin{subfigure}{.33\textwidth}
    \includegraphics[width=1\linewidth]{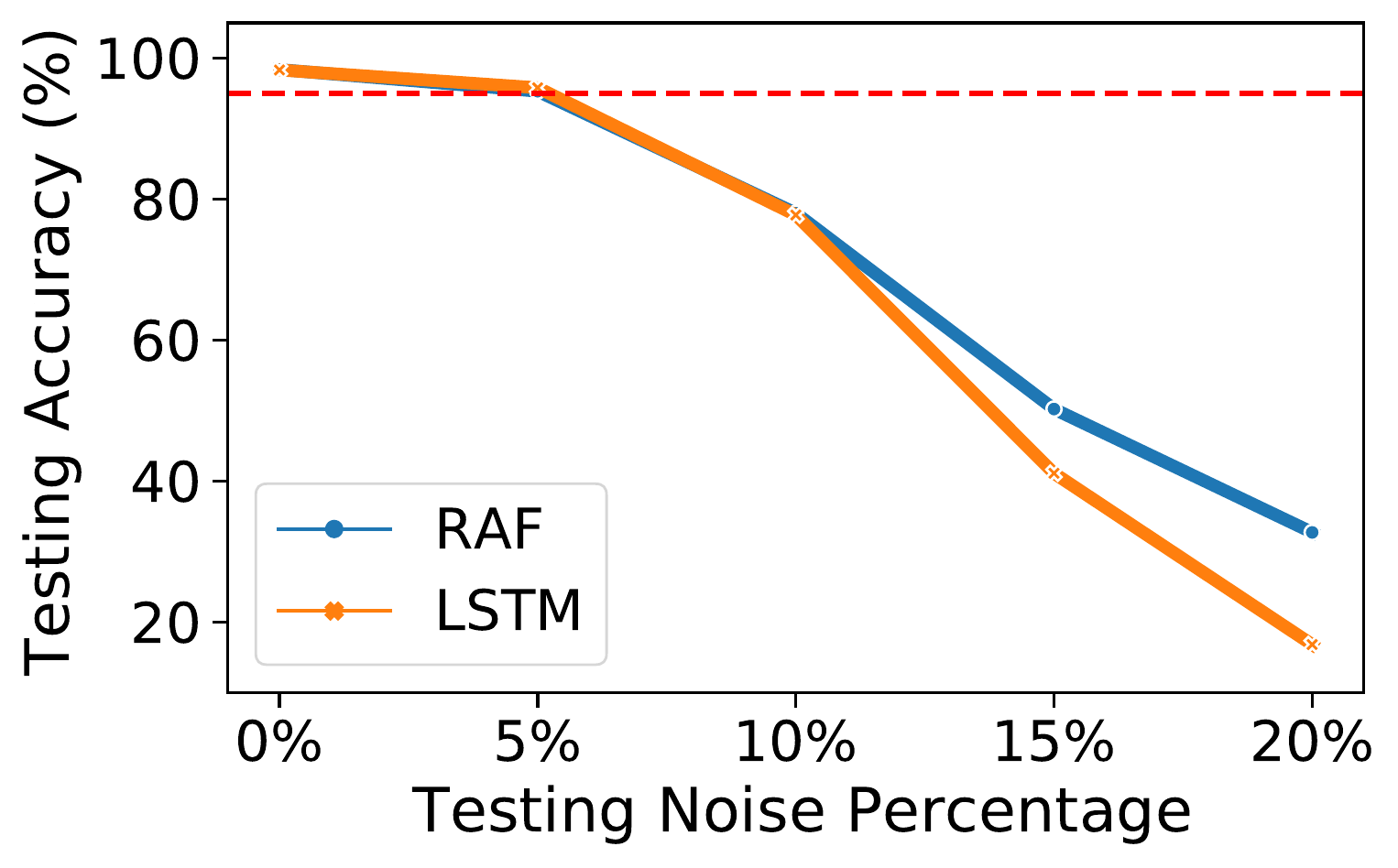}
    \caption{} 
    \label{fig:dynamic_test_noise}
\end{subfigure}%
\begin{subfigure}{.33\textwidth}
    \includegraphics[width=1\linewidth]{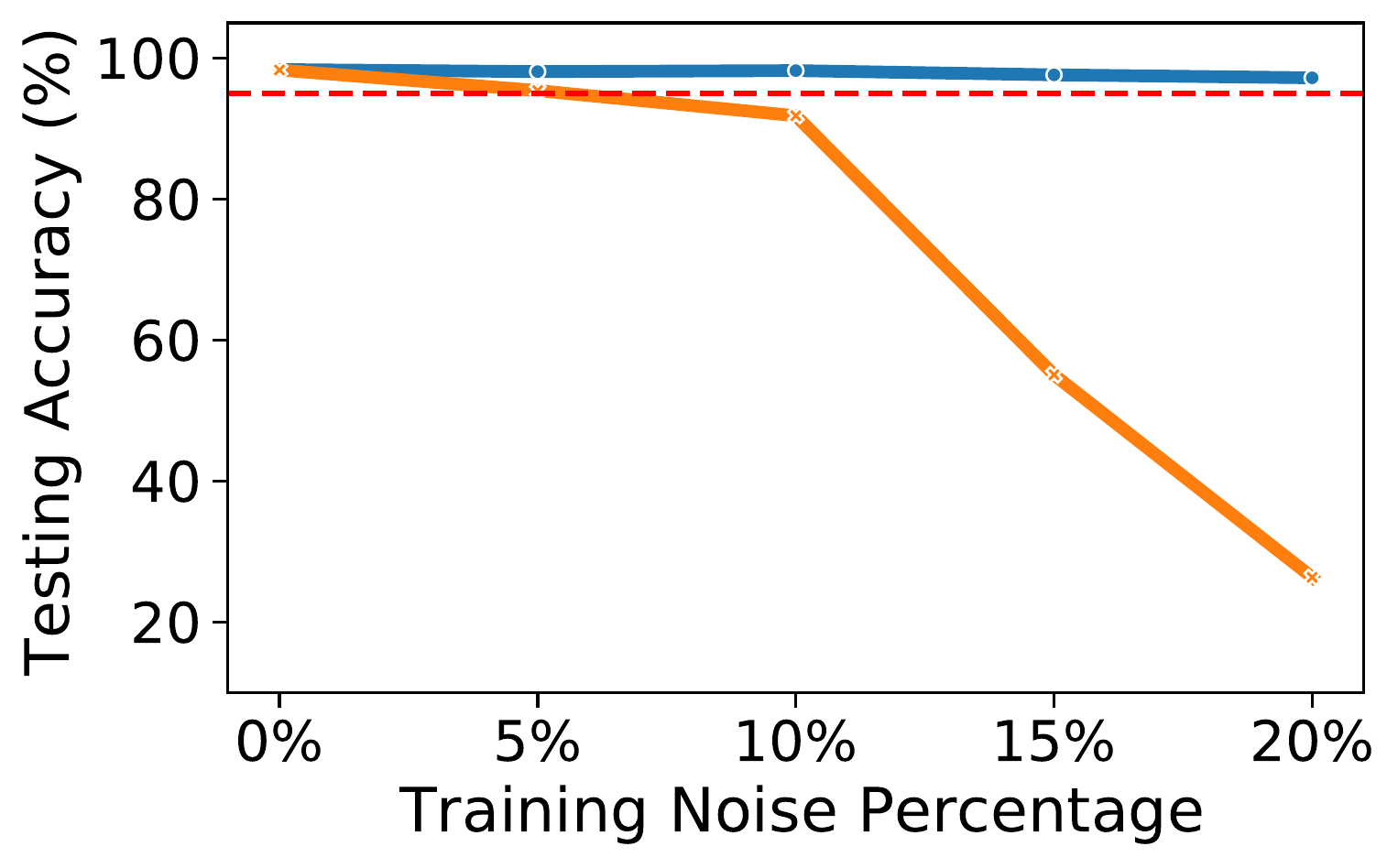}
    \caption{} 
\end{subfigure}%
\begin{subfigure}{.33\textwidth}
    \includegraphics[width=1\linewidth]{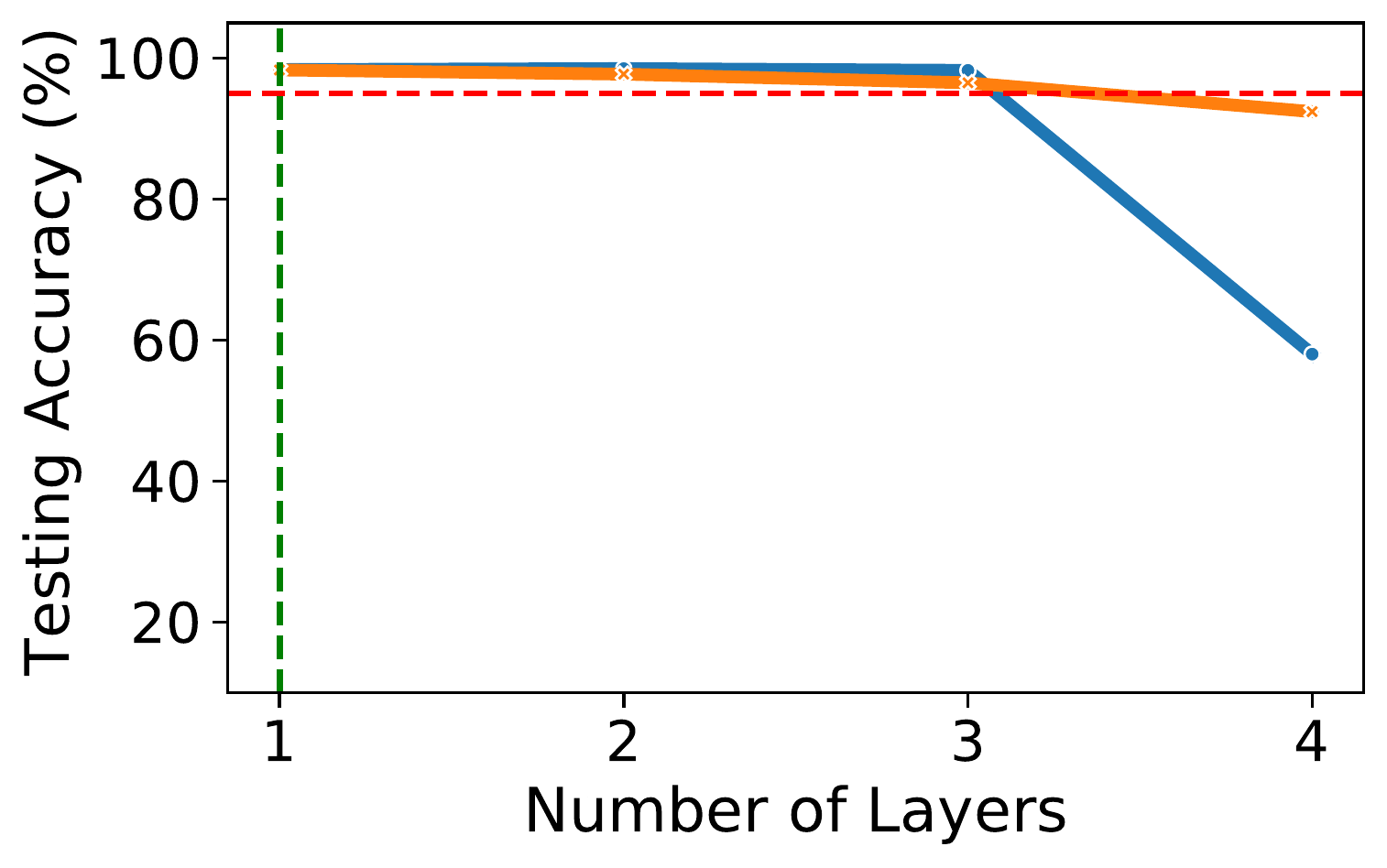}
    \caption{} 
    \label{fig:dynamic_train_noise}
\end{subfigure}
\begin{subfigure}{.33\textwidth}
    \includegraphics[width=1\linewidth]{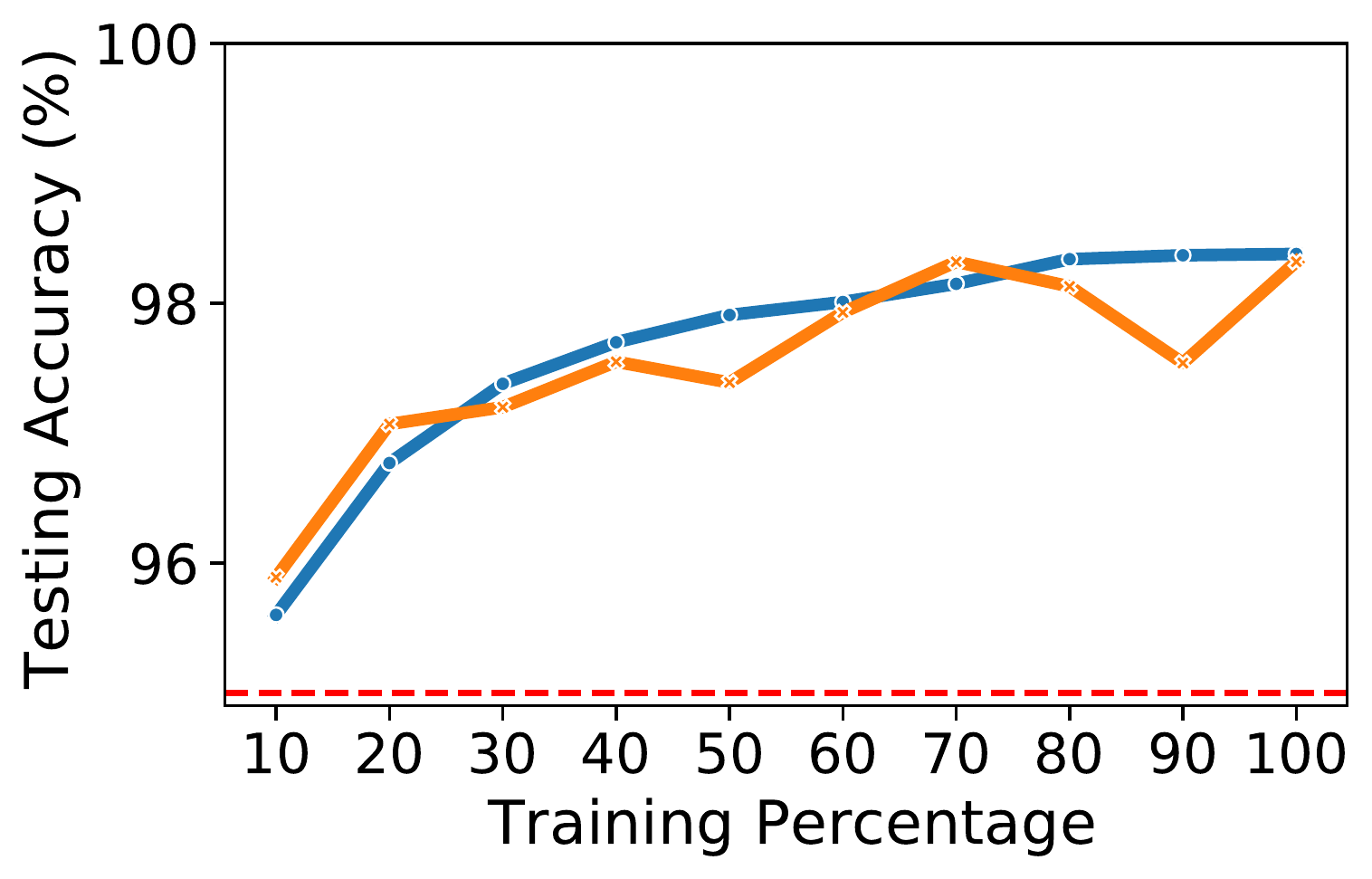}
    \caption{} 
\end{subfigure}%
\begin{subfigure}{.33\textwidth}
    \includegraphics[width=1\linewidth]{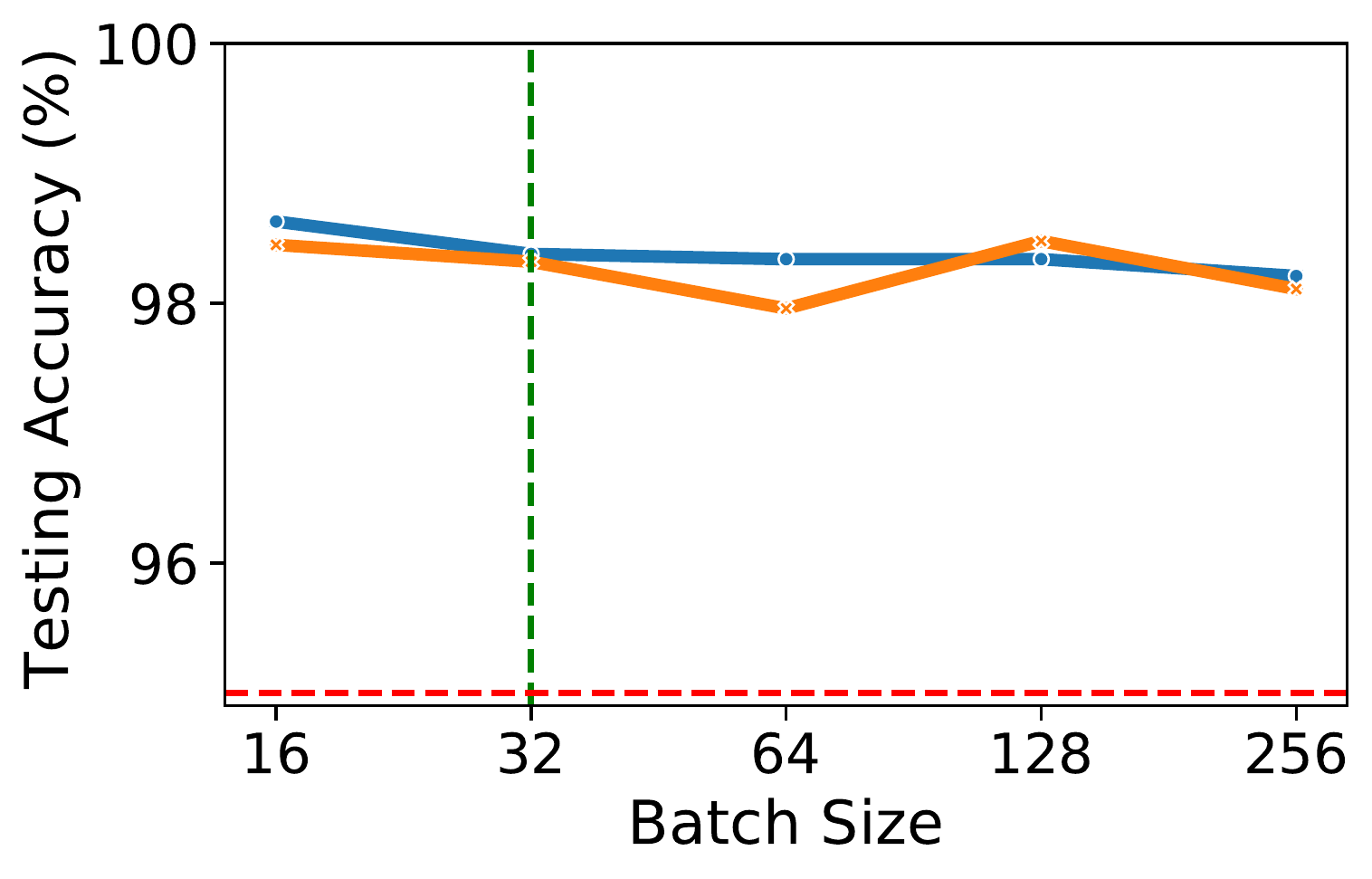}
    \caption{} 
\end{subfigure}%
\begin{subfigure}{.33\textwidth}
    \includegraphics[width=1\linewidth]{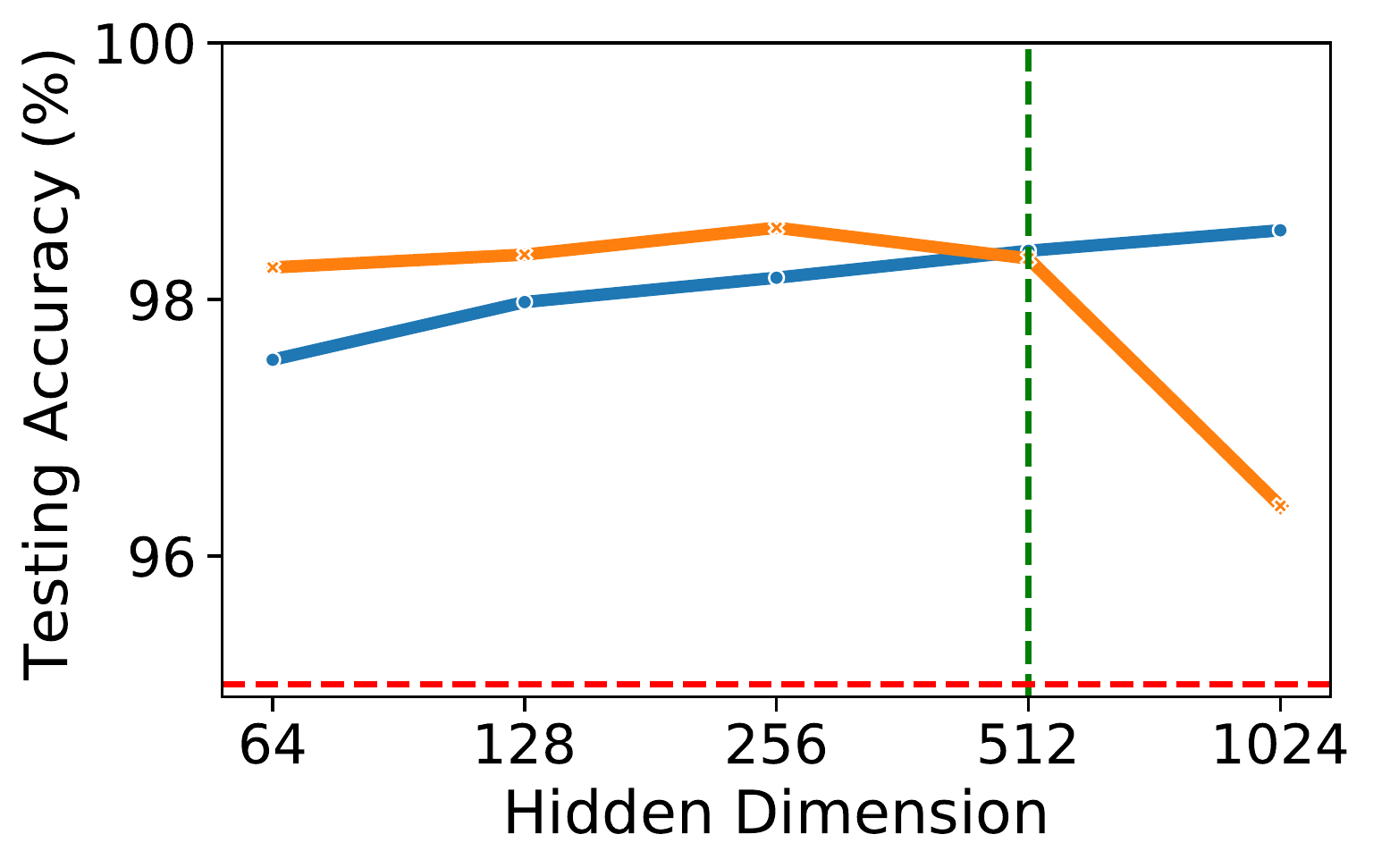}
    \caption{} 
\end{subfigure}
\caption{\textbf{Test accuracy on the N-MNIST dataset (dynamic).}
Comparison of RAF vs. LSTM-networks as a function of: \textbf{(a)} Noise $p$ at test, \textbf{(b)} Noise $p$ at train, \textbf{(c)} Network depth\protect\footnotemark, \textbf{(d)} Training set size, \textbf{(e)} Batch size\footbaseline, and \textbf{(f)} Network width\footbaseline. The red line is a visual anchor at $95\%$ accuracy.
}
\label{fig:dynamic_figures}
\end{figure}

In the dynamic (or neuromorphic) case (Figure \ref{fig:dynamic_figures}), both the RAF and LSTM degrade as we add more noise at \emph{testing} time to networks trained with clean input. At $p\!=\!0.2$, The RAF network gets $32.73\%$ accuracy, twice that of the LSTM. When adding noise to the \emph{training} data and testing on the clean test set, the RAF maintains performance, while the LSTM degrades to nearly $1/4$th of its baseline. As in the static case, performance remains largely comparable between the two models across a range of batch sizes, training set sizes, number of layers, and number of hidden units per layer. Noteworthy are the sudden dips for RAF at $4$ hidden layers, and for LSTM at $1024$ hidden units. See Appendix \ref{app:arch} for further details on all four models. 


\footnotetext{\label{foot:graph_baseline_indicator}The vertical green lines indicate the baseline model used to generate each plot with one varying parameter.}

\section{Conclusions}

In this paper, we presented a novel SNN implementation with certain advantages over conventional ML models: 
    \textbf{(a)} It utilizes a more biologically plausible neuron model: the RAF neuron---able to model an extensive repertoire of experimental observations \citep{Izhikevich_2001},
    \textbf{(b)} It achieves performance similar to homologous deep networks across a range of batch sizes, training sizes, and network widths and depths, with only $\textbf{23.8\%}$ of the competing model size in the dynamic case, and a comparable size in the static case (see Appendix \ref{app:arch}), and 
    \textbf{(c)} It copes more robustly with high noise levels in train/test time compared to homologous approaches in both static and dynamic scenarios (see Figures \ref{fig:static_test_noise}, \ref{fig:dynamic_test_noise} \& \ref{fig:dynamic_train_noise}).
In conclusion, with this approach, we propose a novel SNN that can outperform standard methods, especially in noisy environments. Importantly, it is an interpretable SNN in terms of neuronal parameters, with the long-term goal of using it in neuroscience research. We are also interested in investigating the robustness of RAF networks to adversarial attacks, which is a major shortcoming of conventional ML models.




\bibliography{references}
\bibliographystyle{iclr2021_conference}

\appendix
\section{Appendix}

\subsection{Architecture}
\label{app:arch}

\begin{figure}[h]
    \centering
    \includegraphics[width=0.95\linewidth]{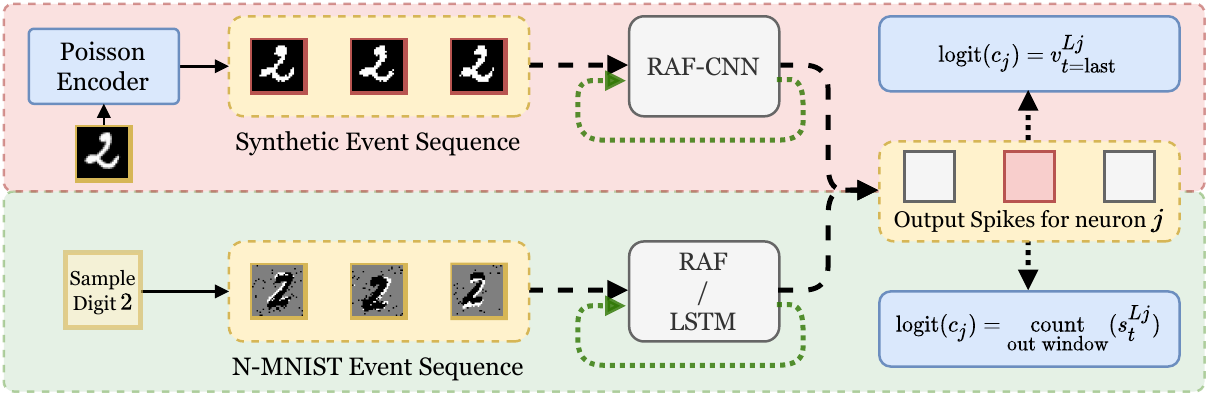}
    \caption{\textbf{Above (red)}: Encoding and model pipeline for the RAF-CNN, trained on MNIST data. \textbf{Below (green):} Model pipeline for the N-MNIST-based experiments with RAF and LSTM networks.}
    \label{fig:pipeline}
\end{figure}

We consider $4$ different models: RAF vs. LSTM, and RAF-CNN vs. CNN. A plain RAF network is analogous to a fully-connected (FC) network with one hidden layer but uses RAF neurons instead and is therefore recurrent. This is compared to an LSTM \citep{lstm} with a similar configuration. Table \ref{tab:raf_lstm} shows the baseline architecture for each along with the number of trainable parameters. Note that the RAF model contains significantly fewer parameters than its counterpart yet outperforms it as discussed in Section \ref{sec:results}.

\begin{table}[H]
\centering
\caption{Comparing the RAF vs. LSTM models}
\begin{tabular}{cc|cc}
\hline
\multicolumn{2}{c|}{\textbf{RAF}}          & \multicolumn{2}{c}{\textbf{LSTM}}          \\ \hline
\textbf{Layer Type}   & \textbf{Dimension} & \textbf{Layer Type}   & \textbf{Dimension} \\ \hline
Input                 & 34x34x2            & Input                 & 34x34x2            \\
Flatten               & 2312               & Flatten               & 2312               \\
RAF                   & 512                & LSTM                  & 512                \\
Linear                & 10                 & Linear                & 10                 \\ \hline
\textbf{\# of Params} & 297,768            & \textbf{\# of Params} & 1,251,594          \\ \hline
\end{tabular}
\label{tab:raf_lstm}
\end{table}

Table \ref{tab:raf_cnn} compares the RAF-CNN model with its non-spiking counterpart. They both follow LeNet architecture as in \cite{lee20enabling}. The RAF-CNN employs convolutional layers in place of FC layers in the RAF. The ``Spiking Average Pooling" layer follows the same spatial-pooling method as in \cite{lee20enabling} but uses the HardSoft activation function \citep{yanguas2020coarse} instead as described in Section \ref{sec:experimental_setup} with the $\beta$ and thresholds made learnable. The ``Convolution" layer in the CNN uses a ReLU activation function. The RAF-CNN contains more trainable parameters because each neuron has a set of parameters: the damping factor ($\xi$), the natural frequency ($\omega$), the firing threshold ($\theta$), and the steepness of the sigmoid ($\beta$)---used to approximate the gradient in the backward pass.

\begin{table}[h]
\centering
\caption{Comparing the RAF-CNN vs. CNN models}
\resizebox{\textwidth}{!}{
\begin{tabular}{cccc|cccc}
\hline
\multicolumn{4}{c|}{\textbf{RAF-CNN}}                                                   & \multicolumn{4}{c}{\textbf{CNN}}                                                        \\ \hline
\textbf{Layer Type}       & \textbf{Kernel Size} & \textbf{Dimension} & \textbf{Stride} & \textbf{Layer Type}       & \textbf{Kernel Size} & \textbf{Dimension} & \textbf{Stride} \\ \hline
Input                     &                      & 28x28x1            &                 & Input                     &                      & 28x28x1            &                 \\
RAF Convolution           & 5x5                  & 128                & 1               & Convolution               & 5x5                  & 128                & 1               \\
Spiking Avg Pooling       & 2x2                  &                    & 2               & Avg Pooling               & 2x2                  &                    & 2               \\
RAF Convolution           & 5x5                  & 128                & 1               & Convolution               & 5x5                  & 128                & 1               \\
Spiking Avg Pooling       & 2x2                  &                    & 2               & Avg Pooling               & 2x2                  &                    & 2               \\
Flatten                   &                      & 6272               &                 & Flatten                   &                      & 6272               &                 \\
Spiking Linear            &                      & 200                &                 & Linear + ReLU             &                      & 200                &                 \\
Spiking Linear            &                      & 10                 &                 & Linear                    &                      & 10                 &                 \\ \hline
\textbf{\# of Parameters} & \multicolumn{3}{c|}{1,671,576}                              & \textbf{\# of Parameters} & \multicolumn{3}{c}{1,669,200}                               \\ \hline
\end{tabular}
}
\label{tab:raf_cnn}
\end{table}

\end{document}